\documentclass[siggraph]{acmart}

\usepackage{booktabs} 
\usepackage{amsmath}
\usepackage{multirow}
\usepackage{mathtools}
\usepackage{pdfpages}

\usepackage{float}
\floatstyle{plaintop}
\restylefloat{table}

\begin{document}
\title{Neuroevolution of Self-Interpretable Agents}


\author{Yujin Tang}
\affiliation{
\institution{Google Brain, Tokyo}
}
\email{yujintang@google.com}

\author{Duong Nguyen}
\affiliation{
\institution{Google Japan}
}
\email{duongnt@google.com}

\author{David Ha}
\affiliation{
\institution{Google Brain, Tokyo}
}
\email{hadavid@google.com}

\begin{abstract}

Inattentional blindness is the psychological phenomenon that causes one to miss things in plain sight.
It is a consequence of the selective attention in perception that lets us remain focused on important parts of our world without distraction from irrelevant details.
Motivated by selective attention, we study the properties of artificial agents that perceive the world through the lens of a \textit{self-attention} bottleneck. By constraining access to only a small fraction of the visual input, we show that their policies are directly interpretable in pixel space.
We find neuroevolution ideal for training self-attention architectures for vision-based reinforcement learning (RL) tasks, allowing us to incorporate modules that can include discrete, non-differentiable operations which are useful for our agent.
We argue that self-attention has similar properties as \textit{indirect encoding}, in the sense that large implicit weight matrices are generated from a small number of key-query parameters, thus enabling our agent to solve challenging vision based tasks with at least 1000x fewer parameters than existing methods.
Since our agent attends to only task critical visual hints, they are able to generalize to environments where task irrelevant elements are modified while conventional methods fail.
\footnote{Videos of our results and source code available at \textbf{\url{https://attentionagent.github.io/}}}

\end{abstract}

%
%

\maketitle

\section{Introduction}

While visual inputs contain rich information, humans are able to quickly locate several task related spots, extract key information and reason about them to take actions~\cite{core_knowledge, brainfacts}.
In the field of machine learning (ML), existing literature has demonstrated successful applications of deep reinforcement learning (RL) to challenging tasks with visual inputs, however, it is unclear whether these agents reason similarly as we do.
This lack of interpretability is one of the major concerns that caused debates in the wider adoption of ML in safety and security prioritized applications.
To get to know what these agents are ``thinking,'' some methods relied on dedicated network architectures and/or carefully designed training schemes \cite{Sorokin2015DeepAR,adebayo2018sanity,DBLP:conf/nips/MottZCWR19}.
Is it possible to build a simple mechanism and design an agent that behaves similarly to the way humans do?

\begin{figure}[t]
\vskip -0.0in
\includegraphics[width=0.4775\textwidth]{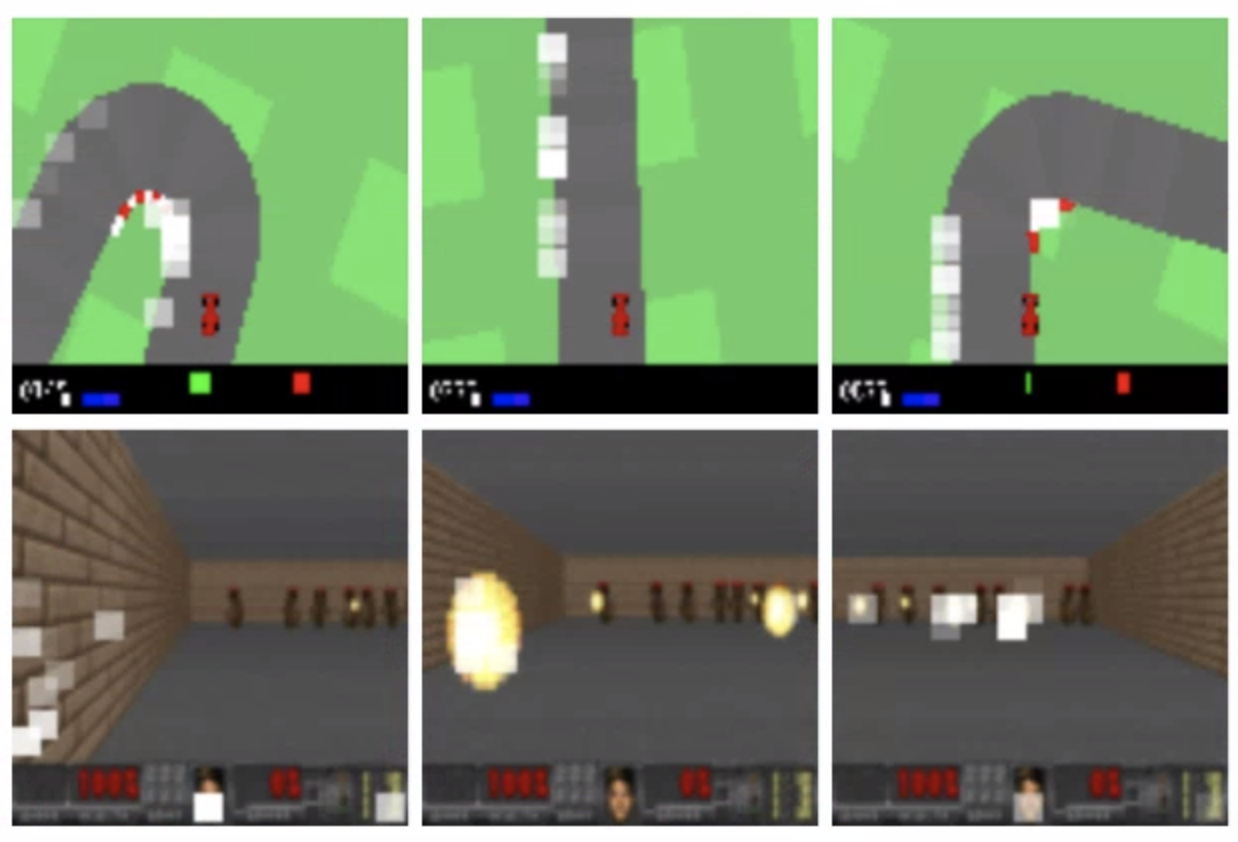}
\vskip -0.15in
\caption{\normalfont In this work, we evolve agents that attend to a small fraction of its visual input critical for its survival, allowing for interpretable agents that are not only compact, but also more generalizable.
Here, we show examples of our agent's attention highlighted in white patches.
In CarRacing (top), our agent mostly attends to the road borders, but shifts its focus to the turns before it changes heading directions. In DoomTakeCover (bottom), the agent is able to focus on fireballs and monsters, consistent with our intuitions.}
\label{fig:demo_figure}
\vskip -0.3in
\end{figure}

One possible solution is to design agents that can encode and process abstract concepts. Recent works have demonstrated the importance of learning abstract information from visual inputs.
For example, the \textit{world models} line of work \cite{ha2018worldmodels,hafner2018learning,kaiser2019model,freeman2019learning} learns compact representations of input image sequences, and have shown that agents trained to use these representations do better in various vision-based tasks.
However, not all elements of an input image are equally important.
As is pointed out recently~\cite{DBLP:journals/corr/abs-1911-08265,gelada2019deepmdp}, an agent does not have to learn representations that are capable of reconstructing the full observation---it is sufficient if the representation allows predicting quantities that are directly relevant for planning. Addressing this, recent works by Risi and Stanley \cite{DBLP:conf/gecco/RisiS19,DBLP:journals/corr/abs-2001-01683} have demonstrated that neuroevolution can be leveraged to train an agent's world model alongside its controller together, end-to-end, even when there are millions of model parameters in the agent's architecture. They also found that the abstract representations learned end-to-end are more task-specific. All these suggests a design of agents that focuses more on and learns from task critical areas of the input.

Most current methods used to train neural networks, whether with gradient descent or evolution strategies, aim to solve for the value of each individual weight parameter of a given neural network.
We refer to these methods as \textit{direct encoding} methods.
\textit{Indirect encoding}~\cite{stanley2003taxonomy,schmidhuber1997discovering}, on the other hand, offers a radically different approach.
These methods optimize instead for a \textit{small} set of rules or operations, referred to as the \textit{genotype}, that specify how the (much larger) neural network (the \textit{phenotype}) should be generated.

Before the popularity of Deep RL, indirect encoding methods in the neuroevolution literature have also been a promising approach for vision-based RL problems. 
For example, earlier works demonstrated that large neural networks can be encoded into much smaller, genotype solutions, that are capable of playing Atari~\cite{hausknecht2012hyperneat} (when it was still considered challenging in 2012) or car racing directly from pixel-only inputs~\cite{koutnik2013evolving}, hinting at its potential power.

By encoding the weights of a large model with a small number of parameters, we can substantially reduce the search space of the solution, at the expense of restricting our solution to a small subspace of all possible solutions offered by direct encoding methods. This constraint naturally incorporates into our agent an \textit{inductive bias} that determines what it does well at~\cite{wann2019,zador2019critique,hasson2020direct}, and this bias is dependent on the choice of our indirect encoding method. For instance, HyperNEAT~\cite{stanley2009hyperneat} has been successful at robotic gait control~\cite{clune2009evolving,risi2013confronting,clune2011performance}, suggesting CPPNs~\cite{stanley2007cppn} to be effective at representing modular and symmetric properties suitable for locomotion. But are better there other encoding methods for vision-based tasks?

In this work, we establish that \textit{self-attention} can be viewed as a form of indirect encoding, which enables us to construct highly parameter-efficient agents.
We investigate the performance and generalization properties of these agents for vision-based RL tasks.
Self-attention has been popularized by Transformer~\cite{NIPS2017_7181} models that have been successfully applied in domains such as natural language processing~\cite{devlin-etal-2019-bert, radford2019language} and vision~\cite{DBLP:journals/corr/abs-1911-03584,DBLP:conf/nips/ParmarRVBLS19,hu2019local,bello2019attention}.
As we will explain, self-attention offers a simple yet powerful approach for parameterizing a large weight matrix of size $\mathcal{O}(n^2)$ using only $\mathcal{O}(d)$ number of parameter values, where $n$ is the size of the visual input, $d$ is the dimension of some transformed space and $n \gg d$.
Furthermore, such a parameterization enforces an inductive bias to encourage our agent to attend to only a small fraction of its visual input, and as such naturally makes the agent more interpretable.



As we will show, neuroevolution is an ideal method for training self-attention agents, because not only can we remove unnecessary complexity required for gradient-based methods, resulting in much simpler architectures, we can also incorporate modules that enhance the effectiveness of self-attention that need not be differentiable. We showcase self-attention agents trained with neuroevolution that require 1000x fewer parameters than conventional methods and yet is able to solve challenging vision-based RL tasks.
Specifically, with less than 4K parameters, our self-attention agents can reach average scores of 914 over 100 consecutive trials in a 2D car racing task~\cite{brockman2016openai} and 1125 in a 3D VizDoom task~\cite{DBLP:journals/tciaig/WydmuchKJ19} (the tasks are considered solved for scores $>900$ and $>750$), comparable with existing state-of-the-art results~\cite{ha2018worldmodels,DBLP:conf/gecco/RisiS19,DBLP:journals/corr/abs-2001-01683}. Moreover, our agent learns to attend to only task critical visual spots and is therefore able to generalize to environments where task irrelevant elements are modified whereas conventional methods fail.

The goal of this work is to showcase self-attention as a powerful tool for the neuroevolution toolbox, and we will open-source code for reproducing our experiments. We hope our results will encourage further investigation into the neuroevolution of self-attention models, and also revitalize interest in indirect encoding methods.

\section{Related Work}

Our work has connections to work in various areas:

\textbf{Neuroscience}\: Although the human visual processing mechanisms are not yet completely understood, recent findings from anatomical and physiological studies in monkeys suggest that visual signals are fed into processing systems to extract high level concepts such as shape, color and spatial organization \cite{freeman2011metamers,brainfacts}.
Research in consciousness suggests that our brains interpret our surrounding environment in a ``language of thought'' that is abstract enough to interface with decision making mechanisms~\cite{dehaene2014consciousness}.
On the other hand, while recent works in deep RL for vision-based tasks have thrived \cite{mnih2013playing,mnih2015humanlevel}, in most cases it is not clear why they~work~or~fail.



%

A line of work that narrows the gap is \textit{world models}~\cite{ha2018worldmodels,hafner2018learning}, where the controller's inputs are abstract representations of the visual and temporal information, provided by encouraging a probabilistic ``world model'' to compress and predict potential future experiences.
Their agent excels in challenging vision-based RL tasks, and by projecting the abstract representation back to the pixel space, it is possible to get some insights into the agent's mind. \cite{DBLP:journals/corr/abs-1911-08265,gelada2019deepmdp} suggest that not all details in the visual input are equally important, specifically they pointed out that rather than learning abstract representations that are capable of reconstructing the full observation, it is sufficient if the representation allows predicting quantities that are directly relevant for planning.

While we do not fully understand the mechanisms of how our brains develop abstract representations of the world, it is believed that attention is the unconscious mechanism by which we can only attend to a few selected senses at a time, allowing our consciousness to condense sensory information into a synthetic code that is compact enough to be carried forward in time for decision making~\cite{mack1998inattentional,vul2009attention,dehaene2014consciousness}. In this work, in place of a probabilistic world model, we investigate the use of self-attention to distill an agent's visual input into small synthetic features as inputs for a small controller.

\textbf{Neuroevolution}-based methods for tackling challenging RL tasks have recently gained popularity due to their simplicity and competitiveness to Deep RL methods, even on vision-based RL benchmark tasks~\cite{salimans2017evolution,ha2017evolving,such2017deep,ha2018designrl,mania2018simple}. More recent work~\cite{DBLP:conf/gecco/RisiS19,DBLP:journals/corr/abs-2001-01683} demonstrated that evolution can even train RL agents with millions of weight parameters, such as the aforementioned world models-based agents. Because these approaches do not require gradient-based computation, they offer more flexibility such as discrete latent codes, being able to optimize directly for the total reward across multiple rollouts, or ease of scaling computation across machines.

It is worthy to note that even before the popularity of deep RL-based approaches for vision-based tasks, indirect encoding methods from the neuroevolution literature have been used to tackle challenging vision-based tasks such as Atari domain~\cite{hausknecht2012hyperneat} and car navigation from pixels~\cite{koutnik2013evolving}. Indirect encoding methods are inspired by biological genotype--phenotype representations, and aim to represent a large but expressive neural network with a small \textit{genotype} code, reducing a high dimensional optimization problem to a more manageable one that can be solved with gradient-free methods.

Indirect encoding methods are not confined to neuroevolution. Inspired by earlier works~\cite{stanley2009hyperneat,schmidhuber1993self}, \textit{hypernetworks}~\cite{ha2017hypernetworks} are recurrent neural networks (RNNs) whose weight matrices can change over time, depending on the RNN's input and state. It uses an outer product projection of an embedding vector, allowing a large weight matrix to be modified via a small genotype embedding. As we will show in the next section, self-attention also relies on taking an outer product of input and other parameter vectors to produce a much larger weight matrix. Transformers~\cite{NIPS2017_7181} demonstrated that this type of modified self-attention matrix is a tremendously powerful prior for various language modeling tasks. Here, we investigate the use of self-attention as an indirect encoding mechanism for training agents to perform vision-based RL~tasks~using~neuroevolution.

\textbf{Attention-based RL}\: Inspired by biological vision systems, earlier works formulated the problem of visual attention as an RL problem~\cite{schmidhuber1991learning,mnih2014recurrent,stollenga2014deep,ba2014multiple,cheung2016emergence}.
Recent work~\cite{zambaldi2018deep} incorporated multi-head self-attention to learn representations that encode relational information between feature entities, with these features the learned agent is able to solve a novel navigation and planning task and achieve SOTA results in six out of seven StarCraft II tasks. Because the agent learned relations between entities, it can also generalize to unseen settings during training.
In order to capture the interactions in a system that affects the dynamics, \cite{goyal2019recurrent} proposed to use a group of modified RNNs. 
Self-attention is used to combine their hidden states and inputs. Each member competes for attention at each step, and only the winners can access the input and also other members' states.
This modular mechanism improved generalization on Atari.

Attention is also explicitly targeted for interpretability in RL.
In \cite{Sorokin2015DeepAR}, the authors incorporated soft and hard attention mechanism into the deep recurrent Q-network, and they were able to outperform Deep Q-network~\cite{mnih2013playing} in a subset of the Atari games.
Most recently, \cite{DBLP:conf/nips/MottZCWR19} used a soft, top-down attention mechanism to force the agent to focus on task-relevant information by sequentially querying its view of the environment.
Their agents achieved competitive performance on Atari while being more interpretable.

Although these methods brought exciting results, they need dedicated network architectures and carefully designed training schemes to work in an RL context.
In \cite{Sorokin2015DeepAR}, a hard attention mechanism had to be separately trained because it required sampling, and in \cite{DBLP:conf/nips/MottZCWR19}, a
carefully designed non-trainable basis that encodes spatial locations was needed. Because we are not concerned with gradient-based learning, we are able to chip away at the complexity and get away with using a much simplified version of the Transformer in our self-attention agent. We do not need to use positional encoding or layer normalization components in the Transformer.

The high dimensionality the visual input makes it computationally prohibitive to apply attention directly to individual pixels, and we rather operate on image \textit{patches} (which have lower dimensions) instead. Although not in the context of self-attention, previous work (e.g. \cite{DBLP:conf/aaai/ChoiLZ17,DBLP:journals/corr/abs-1905-02793,Ding2019ImprovingSS,sumbul2019cnn,DBLP:journals/corr/abs-1904-01784,NIPS2019_8359}) segments the visual input and attend to the patches rather than individual pixels.
Our work is similar to \cite{DBLP:conf/aaai/ChoiLZ17}, where the input to the controller their is a vector of patch features weighted by attentions, the dimension of which grows linearly as we increase the number of patches.
However, as our method do not rely on gradient-based learning, we can simply restrict the input to be only the $K$ patches with highest importance. Ordinal measures have been shown to be robust and used in various feature detectors and descriptors~\cite{rosten2005real}. Using gradient-free methods are more desirable in the case of non-differentiable operations because these ordinal measures can be implemented as $argmax$ or top $K$ patch selection, critical for our self-attention agent.
\vspace{-0.05in}
\section{Background on Self-Attention}


\subsection{Self-Attention}
\label{sec:self_attention}
We now give a brief overview of self-attention. Here, we describe a simpler subset of the full Transformer~\cite{NIPS2017_7181} architecture used in this work. In particular, we omit \textit{Value} matrices, \textit{positional encoding}, \textit{multi-head attention} from our method, and opt for the simplest variation that complements neuroevolution methods for our purpose. 
We refer to \cite{bloem2018} for an in-depth overview of the Transformer model.

Let $X \in \mathcal{R}^{n \times d_{in}}$ be an input sequence of $n$ elements (e.g. number of words in a sentence, pixels in an image), each of dimensions $d_{in}$ (e.g. word embedding size, RGB intensities).
Self-attention module calculates an \textit{attention score matrix} and a weighted output:
\vspace{-0.05in}
\begin{align}
A &= \text{softmax}\big{(}\frac{1}{\sqrt{d_{in}}}(X W_k) (X W_q)^\intercal\big{)}
\label{eq_attention}\\
Y &= A X 
\label{eq_output}
\end{align}
where
$W_k, W_q \in \mathcal{R}^{d_{in} \times d}$ are matrices that map the input to components called \textit{Key} and \textit{Query} ($\text{Key} = XW_k, \text{Query} = XW_q$), $d$ is the dimension of the transformed space and is usually a small integer.
Since the average value of the dot product grows with the vector's dimension, each entry in the Key and Query matrices can be disproportionally too large if $d_{in}$ is large. To counter this, the factor $\frac{1}{\sqrt{d_{in}}}$ is used to normalize the inputs.
Applying the $\text{softmax}$\footnote{$\text{softmax}(x_i) = \exp(x_i) / \sum_{k}{\exp(x_k)}$} operation along the rows of the matrix product in Equation~\ref{eq_attention}, we get the attention matrix $A \in \mathcal{R}^{n \times n}$, where each row vector of $A$ sums to $1$.
Thus, each row of output $Y \in \mathcal{R}^{n \times d_{in}}$ can be interpreted as a weighted average of the input $X$ by each row of the matrix.

Self-attention lets us map arbitrary input $X$ to target output $Y$, and this mapping is determined by an attention matrix $A$ parameterized by much smaller Key and Query parameters, which can be trained using machine learning techniques. The self-attention mechanism is at the heart of recent SOTA methods for translation and language modeling~\cite{devlin-etal-2019-bert, radford2019language}, and has now become a common place method for natural language processing domain.

\subsection{Self-Attention for Images}
Although self-attention is broadly applied to sequential data, it is straightforward to adapt it to images.
For images, the input is a tensor $X \in \mathcal{R}^{H \times W \times C}$ where $H$ and $W$ are the height and width of the image, $C$ is the number of image channels (e.g., 3 for RGB, 1 for gray-scale).
If we reshape the image so that it becomes $X \in \mathcal{R}^{n \times d_{in}}$ where $n = H \times W$ and $d_{in} = C$, all the operations defined in Section~\ref{sec:self_attention} are valid and can be readily applied.
In the reshaped $X$, each row represents a pixel and the attentions are between pixels. Notice that the complexity of Equation~\ref{eq_attention} grows quadratically with the number of rows in $X$ due to matrix multiplication, it therefore becomes computationally prohibitive when the input image is large. While down-sampling the image before applying self-attention is a quick fix, it is accompanied with performance trade-off (\cite{DBLP:conf/nips/ParmarRVBLS19,DBLP:journals/corr/abs-1911-03584} propose methods to partially overcome this trade-off). 


Instead of applying operations on individual pixels of the entire input, a popular method for image recognition is to organize the image into patches and take them as inputs as described in previous work (e.g. \cite{DBLP:journals/corr/abs-1905-02793,Ding2019ImprovingSS,DBLP:journals/corr/abs-1904-01784,NIPS2019_8359}).
In our approach, our agent attends to patches of the input rather than individual pixels, and we use a sliding window to crop the input image in our input transformations.
Conceptually, our approach is similar to \textit{Spatial Softmax}~\cite{finn2016deep,suwajanakorn2018discovery,kulkarni2019unsupervised}, which compresses visual inputs into a set of 2D keypoints that are relevant to the task.
This has been shown to work on robot perception tasks, where the keypoints are spatially interpretable.

\begin{figure*}[!htb]
\vskip -0.20in
\includegraphics[scale=0.441]{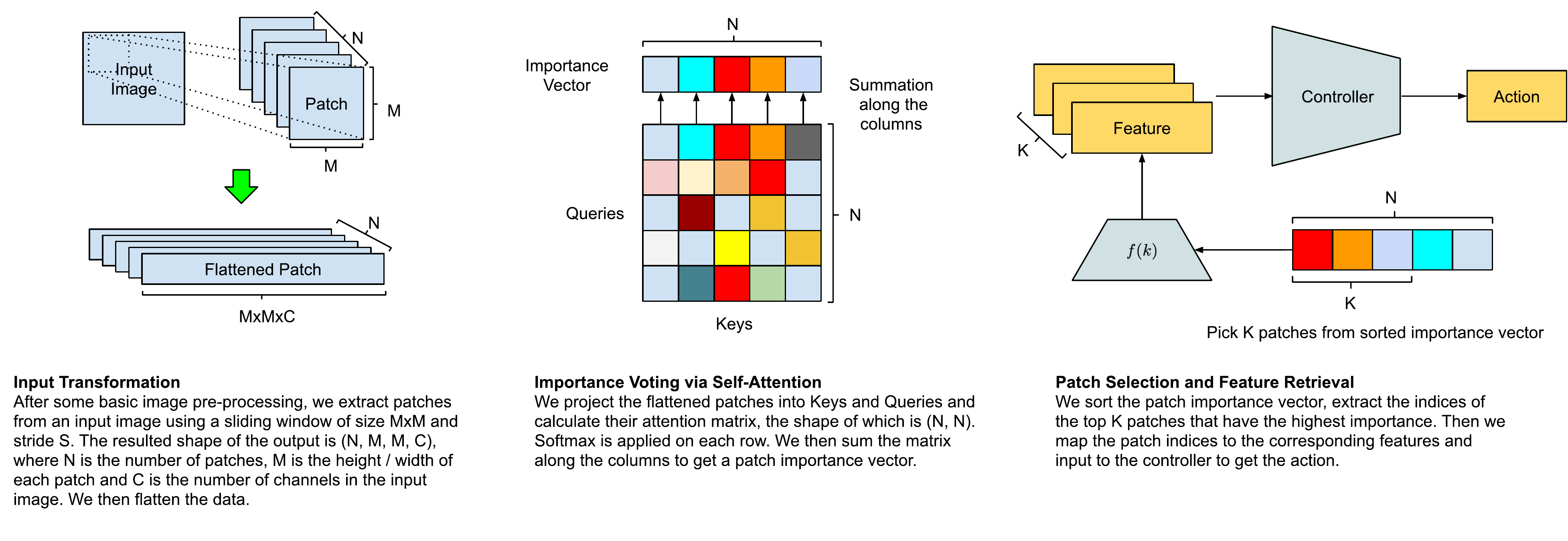}
\vskip -0.275in
\caption{Method overview. \normalfont Illustration of data processing flow in our proposed method.}
\label{fig:algo_flow}
\vskip -0.23in
\end{figure*}

\subsection{Self-Attention as Indirect Encoding}


Indirect encoding methods represent the weights of a neural network, the \textit{phenotype}, with a smaller set of \textit{genotype} parameters. How a genotype encodes a larger solution space is defined by the indirect encoding algorithm. HyperNEAT~\cite{stanley2009hyperneat} encodes the weights of a large network via a coordinate-based CPPN-NEAT~\cite{stanley2007cppn} network, while Compressed Network Search~\cite{koutnik2013evolving} uses discrete cosine transform to compress the weights of a large weight matrix into a small number of DCT coefficients, similar to JPEG compression.

Due to compression, the space of possible weights an indirect encoding scheme can produce is only a small subspace of all possible combination of weights. The constraint on the solution space resulting from indirect encoding enforces an inductive bias into the phenotype. While this inductive bias determines the types of tasks that the network is \textit{naturally} suited at, it also restricts the network to a subset of all possible tasks that an unconstrained phenotype can (in theory) perform. More recent works have proposed ways to broaden its task domain of indirect encoding. \cite{risi2012enhanced} proposed adapting part of the indirect encoding algorithm itself to the task environment. Hypernetworks~\cite{ha2017hypernetworks} suggested making the phenotype directly dependent on the inputs, thus tailoring the weights of the phenotype to the specific inputs of the network. By incorporating information from the input into the weight-generating process, it has been shown~\cite{munkhdalai2017meta,chen2018neural,dumoulin2018featurewise,Oswald2020Continual} that the phenotype can be highly expressive as the weights can adapt to the inputs for the task at hand, while static indirect encoding methods cannot.

Similarly, self-attention enforces a structure on the attention weight matrix $A$ in Equation~\ref{eq_attention} that makes it also input-dependent. If we remove the Key and Query terms, the outer product $X X^T$ defines an association matrix\footnote{$X X^T$ are also known as Gram matrices, and are key to classical statistical learning.} where the elements are large when two distinct input terms are in agreement. This type of structure enforced in $A$ has been shown to be suited for associative tasks where the downstream agent has to learn the relationship between unrelated items. For example, they are used in the Hebbian learning~\cite{hebb1949organization} rule inspired by \textit{neurons that fire together wire together}, shown to be useful for associative learning~\cite{ba2016using,miconi2018differentiable}.
Matrix factorization applied to weights has been proposed in the deep learning literature~\cite{sainath2013low,grosse2016kronecker}, and are also present in recommender systems~\cite{koren2009matrix} to represent relationships between different inputs.

As the outer product $X X^T$ so far has no free parameters, the corresponding matrix $A$ will not be suitable for arbitrary tasks beyond association. The role of the small Key and Query matrices in Equation~\ref{eq_attention} allow $A$ to be modified for the task at hand.
$W_k, W_q \in \mathcal{R}^{d_{in} \times d}$ are the matrices that contain the free parameters, $d_{in}$ is a constant with image inputs (3 for RGB images and 1 for gray scale images), therefore the number of free parameters in self-attention is in the order of $\mathcal{O}(d)$.
As we explained previously, when applying self-attention to images $n$ can be the number of pixels in an input the magnitude of which is often tens of thousands for moderately sized images. On the other hand, $d$ is the dimension of the transformed space in which the Key and Query matrices reside and is often much smaller than $n$ ($d=4$ in our experiments). This form of indirect encoding enables us to represent the phenotype, the attention matrix $A$, of size $\mathcal{O}(n^2)$ using only $\mathcal{O}(d)$ number of genotype parameters. In our experiments, we show that our attention matrix $A$ can be represented using only $\sim$ 1200 trainable genotype parameters.

Furthermore, we demonstrate that features from this attention matrix is especially useful to a downstream decision-making controller. We find that even if we restrict the size of our controller to only $\sim$ 2500 parameters, it can still solve challenging vision-based tasks by leveraging the information provided by self-attention.



\section{Proposed Method}

Encouraged by the success of early Deep RL works such as \cite{mnih2013playing}, most succeeding works adopted variants of earlier models rooted from a similar design. For instance, convolution layers for image down-sampling and feature extraction (the visual feature extraction module) are first used to extract features that are then fed to fully-connected or recurrent layers to produce value estimations or control signals (the controller module).
In such a design, the visual feature extraction module regards each of the elements in the entire input image of equal importance and relies on the training signal to direct its learning so that a small fraction of the weights learn to emphasize task related factors while the others deal with nuances.

Our proposed method is based on a different premise: when the brain is involved in effort-demanding tasks, it assigns most of its attention capacity only to task relevant elements and is temporarily blind to other signals~\cite{mack1998inattentional,kahneman2011thinking}.
In this vein, our agent is designed to focus on only task critical regions in the input image and ignore the others, Figure~\ref{fig:algo_flow} depicts the overview of our proposed method.
To be concrete, given an observation our agent first resizes it into an input image of shape $L \times L$, the agent then segments the image into $N$ patches and regard each patch as a potential region to attend to.
To decide which patches are appropriate, the agent passes the patches to the self-attention module to get a vector representing each patch's importance, based on which it selects $K$ patches of the highest importance.
It then uses the index ($k$) of each of the $K$ patches to fetch relevant features with a function $f(k)$ (described in Sec.~\ref{sec:feature_retrieval}), which can be either a learned module or a pre-defined function that incorporates domain knowledge.
Finally, the agent inputs the features to its controller and generates the action corresponding to the given observation.

To gain a better sense of the magnitudes involved, we summarize the hyper-parameters used in this work in Table~\ref{table:hp}. Some of the parameters are explained in the following sections.

\begin{table}[!htb]
\begin{small}
\vskip -0.05in
\caption{Hyper-parameters in this paper.
\normalfont Left: Parameters for input transformation. After resizing the observation into an image of shape $L \times L$, we use a sliding window of specified size and stride to segment the image into $N = (\left \lfloor \frac{L-M}{S} + 1 \right \rfloor)^2 = (\left \lfloor \frac{96 - 7}{4} + 1 \right \rfloor)^2 = 529$ patches.
Right: Parameters for self-attention. Since the attention is between patches and each patch is RGB, we therefore have $d_{in} = M^2 \times C = 7^2 \times 3 = 147$.}
\label{table:hp}
\vskip -0.00in 
\begin{tabular}{|l|l|l|l|l|}
\cline{1-2} \cline{4-5}
\textbf{Parameter} & \textbf{Value} &  & \textbf{Parameter} & \textbf{Value} \\ \cline{1-2} \cline{4-5} 
Input size ($L$)   & $96$           &  & $d_{in}$           & $147$          \\ \cline{1-2} \cline{4-5} 
Window size ($M$)  & $7$            &  & $d$                & $4$            \\ \cline{1-2} \cline{4-5} 
Stride ($S$)       & $4$            &  & $K$                & $10$          \\ \cline{1-2} \cline{4-5} 
\# of Patches ($N$ and $n$)    & $529$     &  & \# of LSTM neurons  & $16$           \\ \cline{1-2} \cline{4-5}
\end{tabular}
\end{small}
\vskip -0.175in 
\end{table}

\subsection{Input Transformation}
Our agent does some basic image processing and then segments an input image into multiple patches.
For all the experiments in this paper, our agent receives RGB images as its input, therefore we simply divide each pixel by 255 to normalize the data, but it should be straightforward to integrate other data preprocessing procedures.
Similarly, while there can be various methods for image segmentation, we find a simple sliding window strategy to be sufficient for the tasks in this paper.
To be concrete, when the window size $M$ and stride $S$ are specified, our agent chops an input of shape $(H, W, C)$ into a batch of $N$ patches of shape $(M, M, C)$, where $H$ and $W$ are the height and width of the input image and $C$ is the number of channels.
We then reshape the processed data into a matrix of shape $(N, M \times M \times C)$ before feeding it to the self-attention module.
$M$ and $S$ are hyper-parameters to our model that determine how large each patch is and whether patches overlap.
In the extreme case when $M = S = 1$ this becomes self-attention on each individual pixel in the image.

\subsection{Importance Voting via Self-Attention}
Upon receiving the transformed data in $\mathcal{R}^{n \times d_{in}}$ where $n=N$ and $d_{in} = M \times M \times C$, the self-attention module follows Equation~\ref{eq_attention} to get the attention matrix of shape $(N, N)$.
To keep the agent as simple as possible, we do not use positional encoding in this work.

By applying softmax, each row in the attention matrix sums to one, so the attention matrix can be viewed as the results from a voting mechanism between the patches.
To be specific, if each patch can distribute fractions of a total of 1 vote to other patches (including itself), row $i$ thus shows how patch $i$ has voted and column $j$ gives the votes that patch $j$ acquired from others.
In this interpretation, entry $(i, j)$ in the attention matrix is then regarded as how important patch $j$ is from patch $i$'s perspective.
Taking sums along the columns of the attention matrix results in a vector that summarizes the total votes acquired by each patch, and we call this vector the \textit{patch importance vector}.
Unlike conventional self-attention operations, we rely solely on the patch importance vector and do not calculate a weighted output with Equation~\ref{eq_output}.

\subsection{Patch Selection and Feature Retrieval}
\label{sec:feature_retrieval}
Based on the patch importance vector, our agent picks the $K$ patches with the highest importance. We pass in the index of these $K$ patches (denoted as index $k$ to reference the $k^{th}$ patch) into a \textit{feature retrieval operation} $f(k)$ to query the for their features. $f(k)$ can be static mappings or learnable modules, and it returns the features related to the image region centered at patch $k$'s position. The following list gives examples of possible features:
\begin{itemize}
    \item \textbf{Patch center position.} $f(k): \mathcal{R} \mapsto \mathcal{R}^2$ where the output contains the row and column indices of patch $k$'s center.
    \item \textbf{Patch's image histogram.} $f(k): \mathcal{R} \mapsto \mathcal{R}^{b}$ where the output is the image histogram calculated from patch $k$ and $b$ is the number of bins.
    \item \textbf{Convolution layers' output.} $f(k): \mathcal{R} \mapsto \mathcal{R}^{s \times s \times m}$ is a stack of convolution layers (learnable or fixed with pre-trained weights). It takes the image region centered at patch $k$ as input and outputs a tensor of shape $s \times s \times m$.
\end{itemize}
In this work, we use the first example for its simplicity. These design choices give us control over our agent's capabilities and computational efficiency, and will also affect its interpretability.

By discarding patches of low importance the agent becomes temporarily blind to other signals, this is built upon our premise and effectively creates a bottleneck that forces the agent to focus on patches only if they are critical to the task.
Once learned, we can visualize the $K$ patches and see directly what the agent is attending to (see Figure~\ref{fig:demo_figure}).
Although this mechanism introduces $K$ as a hyper-parameter, we find it easy to tune (along with $M$ and $S$). In principle we can also let neuroevolution decide on the number of patches, and we will leave this for future work.

Pruning less important patches also leads to the reduction of input features, so the agent is more efficient by solving tasks with fewer weights.
Furthermore, correlating the feature retrieval operation $f(k)$ with individual patches can also lower the computational cost.
For instance, if some local features are known to be useful for the task yet computationally expensive, $K$ acts as a budget cap and only compute features from the most promising regions.
Notice however, this does not imply we permit only local features, as $f(k)$ also has the flexibility to incorporate global features.
In this paper, $f(k)$ is a simple mapping from patch index to patch position in the image and is a local feature.
But $f(k)$ can also be a stack of convolution layers whose receptive fields are centered at patch $k$. If the receptive fields are large, $f(k)$ can provide global features.

\subsection{Controller}

Temporal information between steps is important to most RL tasks, but single RGB images as our input at each time step do not provide this information.
One option is to stack multiple input frames like what is done in \cite{mnih2015humanlevel}, but we find this inelegant approach unsatisfactory because the time window we care about can vary for different tasks. 
Another option is to incorporate the temporal information as a hidden state inside our controller. Previous work~\cite{cuccu2019playing} has demonstrated that with a good representation of the input image, even a small RNN controller with only 6--18 neurons is sufficient to perform well at several Atari games using only visual inputs.

In our experiments, we use Long short-term memory (LSTM)~\cite{hochreiter1997long} network as our RNN controller so that its hidden state can capture temporal information across multiple input image frames. We find that an LSTM with only 16 neurons is sufficient to solve challenging tasks when combined with features extracted from self-attention.

\subsection{Neuroevolution of the Agent}

Operators such as importance sorting and patch pruning in our proposed methods are not gradient friendly. It is not straightforward to apply back-propagation in the learning phase.
Besides, restricting to gradient based learning methods can prohibit the adoption of learnable feature retrieval functions $f(k)$ that consist of discrete operations or produce discrete features.
We therefore turn to evolution algorithms to train our agent.
While it is possible to train our agent using any evolution strategy or genetic algorithms, empirically we find the performance of Covariance Matrix Adaptation Evolution Strategy (CMA-ES)~\cite{Hansen2006} stable on a set of tasks~\cite{es_on_gke, otoro_blog}.
Despite its power, the computation of the full covariance matrix is non-trivial, and because of this CMA-ES is rarely applied to problems in high-dimensional space \cite{10.1007/978-3-319-99259-4_33} such as the tasks dealing with visual inputs.
As our agent contains significantly fewer parameters than conventional methods, we are therefore able to train it with an off-the-shelf implementation of CMA-ES~\cite{hansen2019pycma}.


\section{Experiments}\label{sec:exp}

Our goal in this section is to answer the following questions via experiments and analysis:
\begin{enumerate}
    \item Is our agent able to solve challenging vision-based RL tasks? If so, what are the advantages over other methods?
    \item How robust is the learned agent? If the agent is focusing on task critical factors, does it generalize to the environments with modifications that are irrelevant to the core mission?
\end{enumerate}

\subsection{Task Description}

\begin{figure}[!htb]
\vskip -0.15in
\includegraphics[scale=0.27]{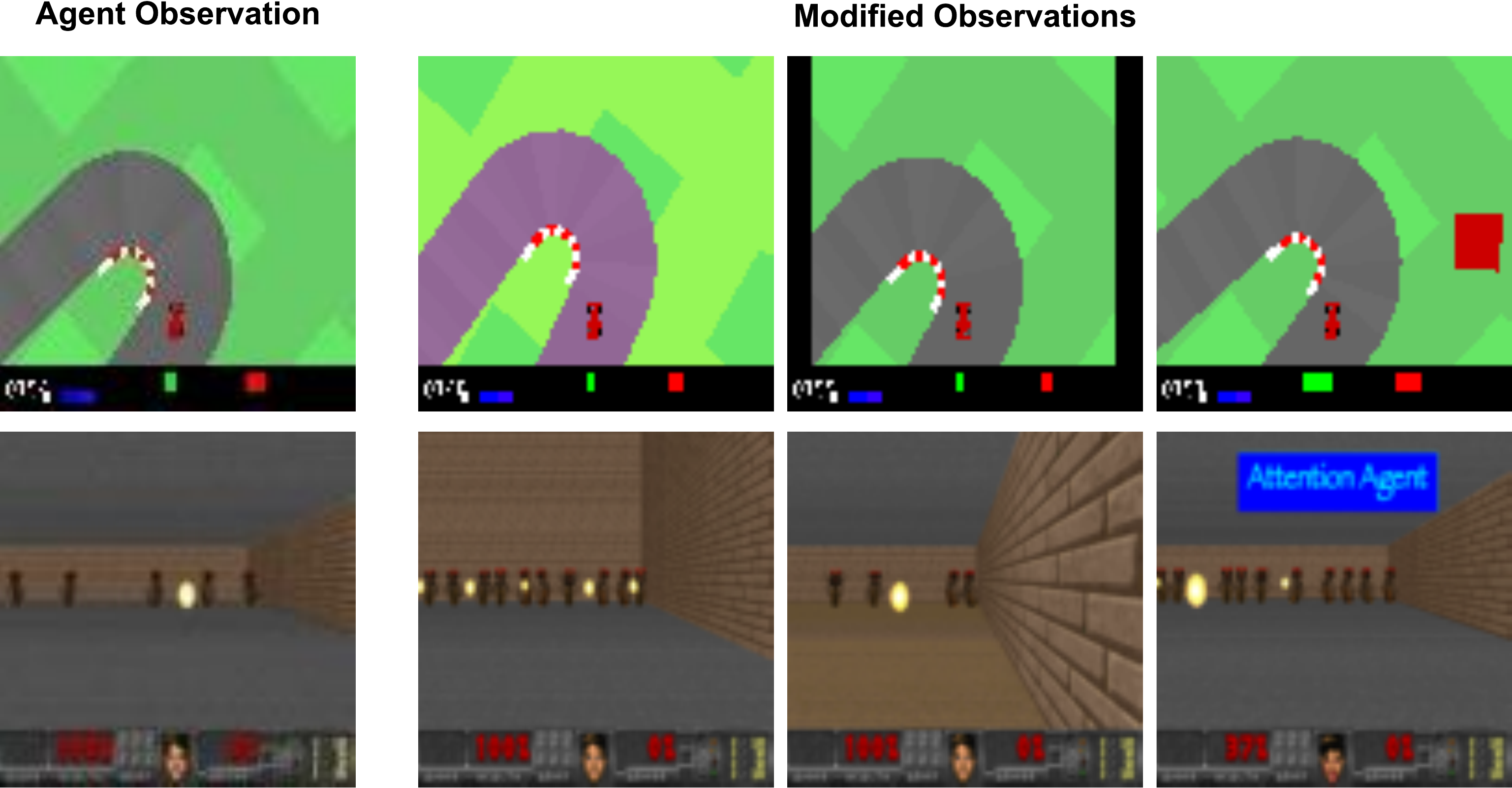}
\vskip -0.1in
\caption{CarRacing and DoomTakeCover. \normalfont \textit{Left:} Original tasks. The observations are resized to 96x96px and presented to the agent. \textit{Right:} Modified CarRacing environments: Color Perturbation, Vertical Frames, Background Blob. Modified DoomTakeCover environments: Higher Walls, Different Floor Texture, Hovering Text.
}
\label{fig:sample_screens}
\vskip -0.15in
\end{figure}
We evaluate our method in two challenging vision-based RL tasks: CarRacing~\cite{CarRacing-v0} and DoomTakeCover~\cite{DBLP:conf/cig/KempkaWRTJ16, DoomTakeCover-v0}. See the first column of Figure~\ref{fig:sample_screens} for sample screenshots from these tasks.

In CarRacing, the agent controls three continuous actions (steering left/right, acceleration and brake) of the red car to visit as many randomly generated track tiles as possible in limited steps.
At each step, the agent receives a penalty of $-0.1$ but will be rewarded with a score of $+\frac{1000}{n}$ for every track tile it visits where $n$ is the total number of tiles.
Each episode ends either when all the track tiles are visited or when 1000 steps have passed.
CarRacing is considered solved if the average score over 100 consecutive test episodes is higher than 900.
Numerous works have tried to tackle this task with Deep RL algorithms,
but it is not solved until recently by~\cite{ha2018worldmodels,DBLP:conf/gecco/RisiS19,DBLP:journals/corr/abs-2001-01683}. For comparison purposes, we also include Proximal Policy Optimization (PPO)~\cite{schulman2017proximal,xtma_ppo}, a popular Deep RL algorithm, as a baseline.

VizDoom~\cite{DBLP:conf/cig/KempkaWRTJ16} serves as a platform for the development of agents that play DOOM using visual information. DoomTakeCover~\cite{DoomTakeCover-v0} is a task in VizDoom where the agent is required to dodge the fireballs launched by the monsters and stay alive for as long as possible.
Each episode lasts for 2100 steps but ends early if the agent dies from being shot.
This is a discrete control problem where the agent can choose to move left/right or stay still at each step.
The agent gets a reward of $+1$ for each step it survives, and the task is regarded solved if the average accumulated reward over 100 episodes is larger than 750.
While a pre-trained world model~\cite{ha2018worldmodels} is able to solve both CarRacing and this task, it has been reported~\cite{DBLP:journals/corr/abs-2001-01683} that the end-to-end direct encoding genetic algorithm (GA) proposed by Risi and Stanley~\cite{DBLP:conf/gecco/RisiS19} falls short at solving this task without incorporating multi-objective optimization~\cite{DBLP:journals/corr/abs-2001-01683} to preserve diversity.

\subsection{Agent Settings}

\begin{figure}[!htb]
\vskip -0.0in
\includegraphics[scale=0.30]{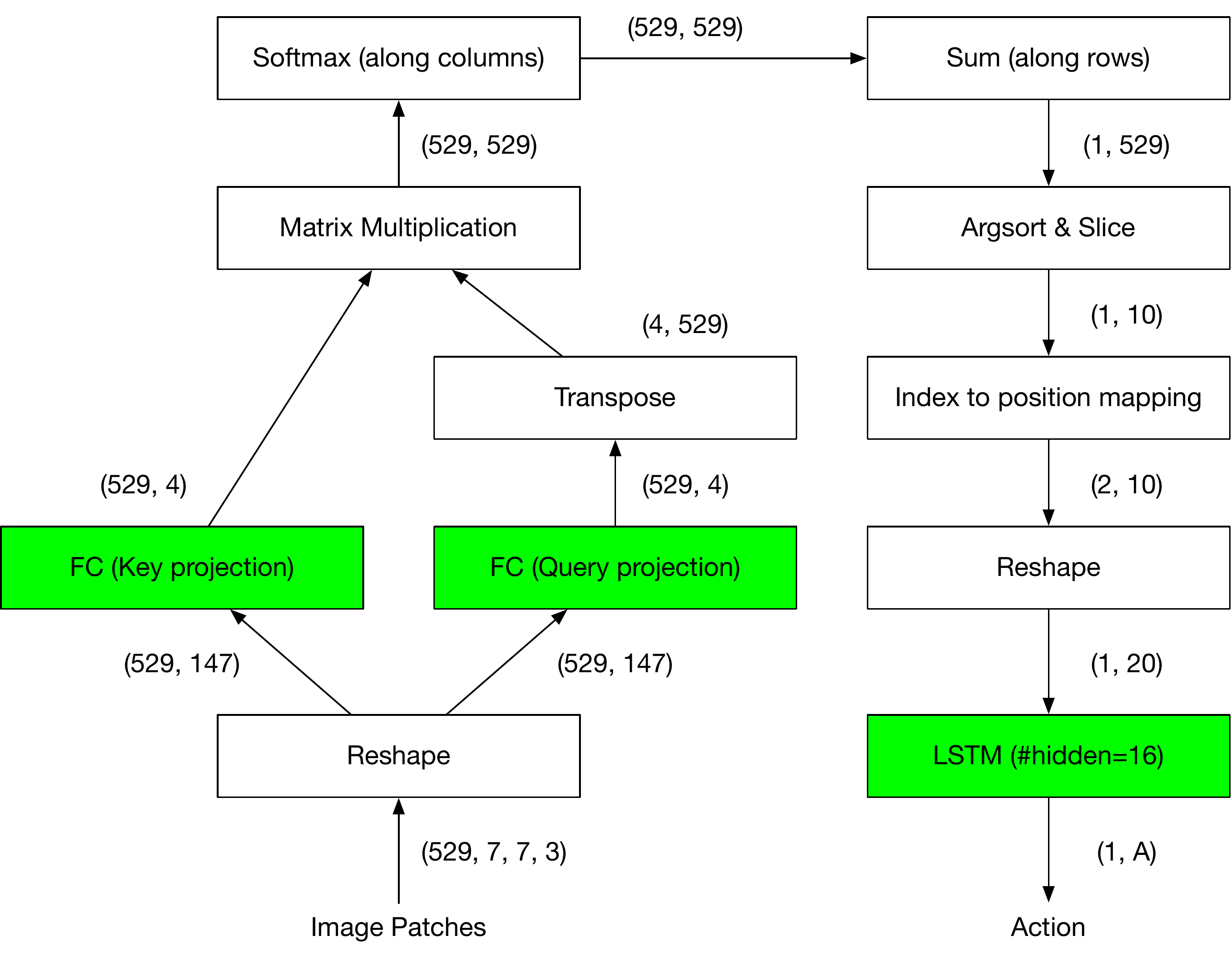}
\vskip -0.1in
\caption{Agent network architecture. \normalfont
Numbers next to each arrow indicate data shape after the operation.
$A$ is the action dimension and is task dependent. Learnable parameters in green.}
\label{fig:agent_arch}
\vskip -0.15in
\end{figure}

Our network architecture and related parameters are shown in Figure~\ref{fig:agent_arch}.
We resize the input images to $96 \times 96$ and use the same architecture for both CarRacing and DoomTakeCover (except for the output dimensions).
We use a sliding window of size $M=7$ and stride $S=4$ to segment the input image, this gives us $N = 529$ patches.
After reshaping, we get an input matrix $X$ of shape $(n=529, d_{in}=147)$.
We project the input matrix to Key and Query with $d=4$, after self-attention is applied we extract features from the $K=10$ most importance patches and input to the single layer LSTM controller (\#hidden=16) to get the action.
Table~\ref{table:num_params} summarizes the number of parameters in our agent, we have also included models from some existing works for the purpose of comparison .
For feature retrieval function $f(k)$, we use a simple mapping from patch index to patch center position in the input image.
We normalize the positions by dividing the largest possible value so that each coordinate is between 0 and 1.

\begin{table}[!htb]
\vskip -0.1in
\begin{small}
\caption{Learnable parameters.
\normalfont GA, DIP share the same world model architecture. Fully connected (FC) layers include bias term.}
\label{table:num_params}
\begin{tabular}{|l|l|l|l|}
\hline
\textbf{Method} & \textbf{ Component} & \textbf{\#Params} & \textbf{Total} \\ \hline
\multirow{3}{*}{\begin{tabular}[c]{@{}l@{}}World model~\cite{ha2018worldmodels}\\
GA~\cite{DBLP:conf/gecco/RisiS19} \\
DIP~\cite{DBLP:journals/corr/abs-2001-01683}\end{tabular}}
& VAE & 4,348,547 & 
\multicolumn{1}{c|}{\multirow{3}{*}{4,733,485}} \\ \cline{2-3} & MD-RNN & 384,071 & 
\multicolumn{1}{c|}{} \\ \cline{2-3} & Controller & 867 & \multicolumn{1}{c|}{} \\ \hline
\multirow{2}{*}{PPO~\cite{xtma_ppo}} & Conv Stack & 393,848 & \multirow{2}{*}{445,955} \\ \cline{2-3}
& FC Stack & 52,107 & \\ \cline{2-3}
\hline
\multirow{3}{*}{Ours} & FC (Query) & 592 & \multirow{3}{*}{3,667} \\ \cline{2-3}
& FC (Key)  & 592  & \\ \cline{2-3}
& LSTM (\#hidden=16) & 2,483 & \\ \hline
\end{tabular}
\end{small}
\vskip -0.1in
\end{table}

We use pycma~\cite{hansen2019pycma}, an off-the-shelf implementation of CMA-ES~\cite{Hansen2006} to train our agent.
We use a population size of 256, set the initial standard deviation to 0.1 and keep all other parameters at default values.
To accurately evaluate the fitness of each individual in the population, we take the mean score over 16 rollouts in CarRacing and 5 rollouts in DoomTakeCover as the its fitness.

\subsection{Results}

\begin{figure}[t]
\vskip -0.0in
\includegraphics[scale=0.27]{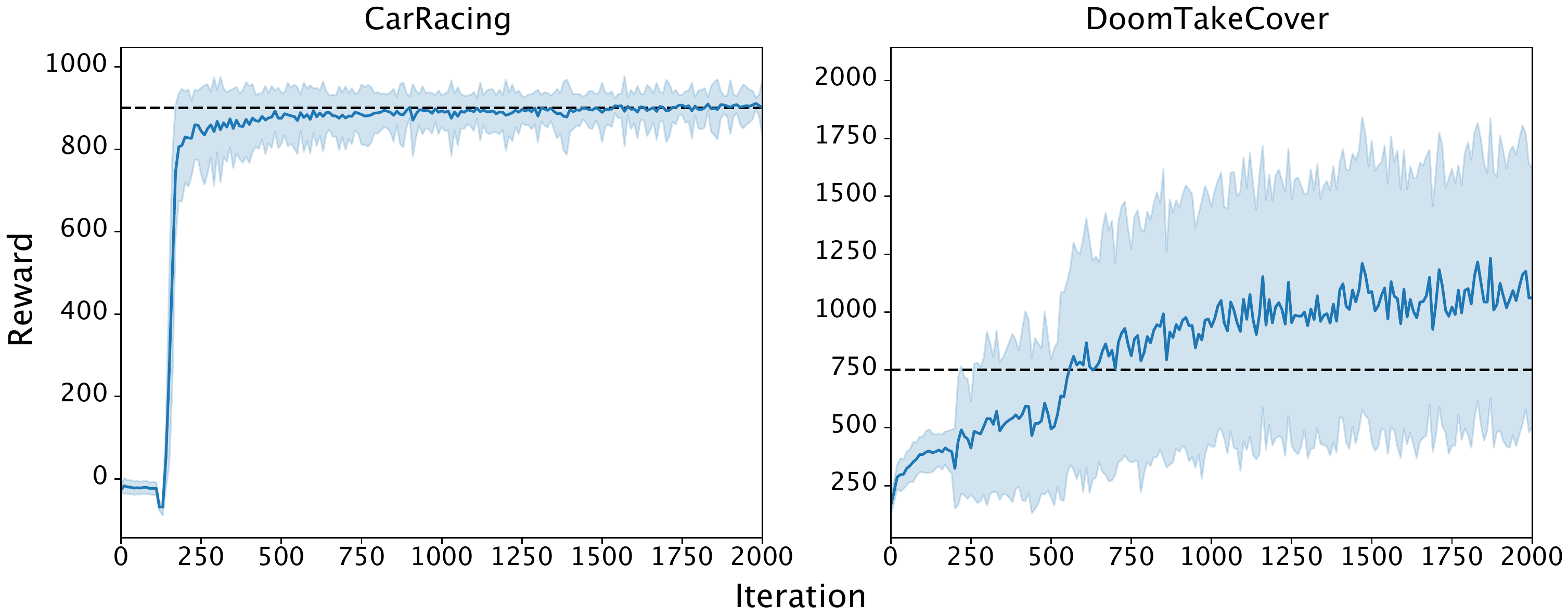}
\vskip -0.1in
\caption{Test scores vs training iterations. \normalfont We test our agent for 100 consecutive episodes with different environment seeds every 10 training iterations. The solid line shows the average score, the shaded area gives the standard deviation and the dashed line indicates the score above which the task is considered solved.}
\label{fig:learning_curve}
\vskip -0.20in
\end{figure}

Not only is our agent able to solve both tasks, it also outperformed the existing methods, Table~\ref{table:scores} summarizes our agent's scores.
From the learning curves in Figure~\ref{fig:learning_curve},
our agent is able to reach the required score within 1000 steps.
Given more training budgets, CMA-ES stably trains our agent to achieve higher scores.

In addition to the SOTA scores, the attention patches visualized in pixel space also make it easier for humans to understand the decisions made by our agent.
In Figure.~\ref{fig:demo_figure}, we plot the top $K$ important patches elected by the self-attention module on top of the input image and see directly what the agent is attending to (see the accompanying videos for more results).
The opacity indicates the importance, the whiter the more important.

\begin{table}[!htb]
\vskip -0.1in
\begin{small}
\caption{Scores from CarRacing and DoomTakeCover.
\normalfont We report the average score over 100 consecutive tests with standard deviations.
For reference, the required scores above which the tasks are considered solved are also included. Best scores are highlighted.}
\label{table:scores}
\begin{tabular}{|l|l|l|}
\hline
\textbf{Method} & \textbf{CarRacing} & \textbf{DoomTakeCover} \\ \hline
Required score  & $900$ & $750$ \\ \hline
World model~\cite{ha2018worldmodels}      & $906 \pm 21$ & $1092 \pm 556$ \\ \hline
PPO~\cite{xtma_ppo} & $865 \pm 159$ & - \\ \hline
GA~\cite{DBLP:conf/gecco/RisiS19}            & $903 \pm 73$ & -              \\ \hline
DIP~\cite{DBLP:journals/corr/abs-2001-01683} & -            & $824 \pm 492$  \\ \hline
Ours                       & $\mathbf{914 \pm 15}$ & $\mathbf{1125 \pm 589}$ \\ \hline
\end{tabular}
\end{small}
\vskip -0.125in
\end{table}

From the figures, we notice that most of the patches the agent attends to are consistent with humans intuition.
For example in CarRacing, the agent's attention is on the border of the road but shifts its focus to the turns before the car needs to change its heading direction.
Notice the attentions are mostly on the left side of the road. This makes sense from a statistical point of view considering that the racing lane forms a closed loop and the car is always running in a counter-clockwise direction.
In DoomTakeCover, the agent is able to focus its attention on fireballs.
When the agent is near the corner of the room, it is also able to detect the wall and change its dodging strategy instead of stuck into the dead end.
Notice the agent also distributes its attention on the panel at the bottom, especially on the profile photo in the middle.
We suspect this is because the controller is using patch positions as its input, and it learned to use these points as anchors to estimate its distances to the fireballs.
We also notice that the scores from all methods have large variance in DoomTakeCover.
This seems to be caused by the environment's design: some fireballs might be out of the agent's sight but are actually approaching. The agent can still be hit by them when it's dodging other fireballs that are in the vision.

Through these tasks, we are able to give a positive answer to the first question in Section~\ref{sec:exp}.
Our agent is indeed able to solve these vision-based RL challenges, and it is efficient in terms of being able to reach higher scores with significantly fewer parameters.

\begin{figure}[htb!]
\includegraphics[scale=0.19]{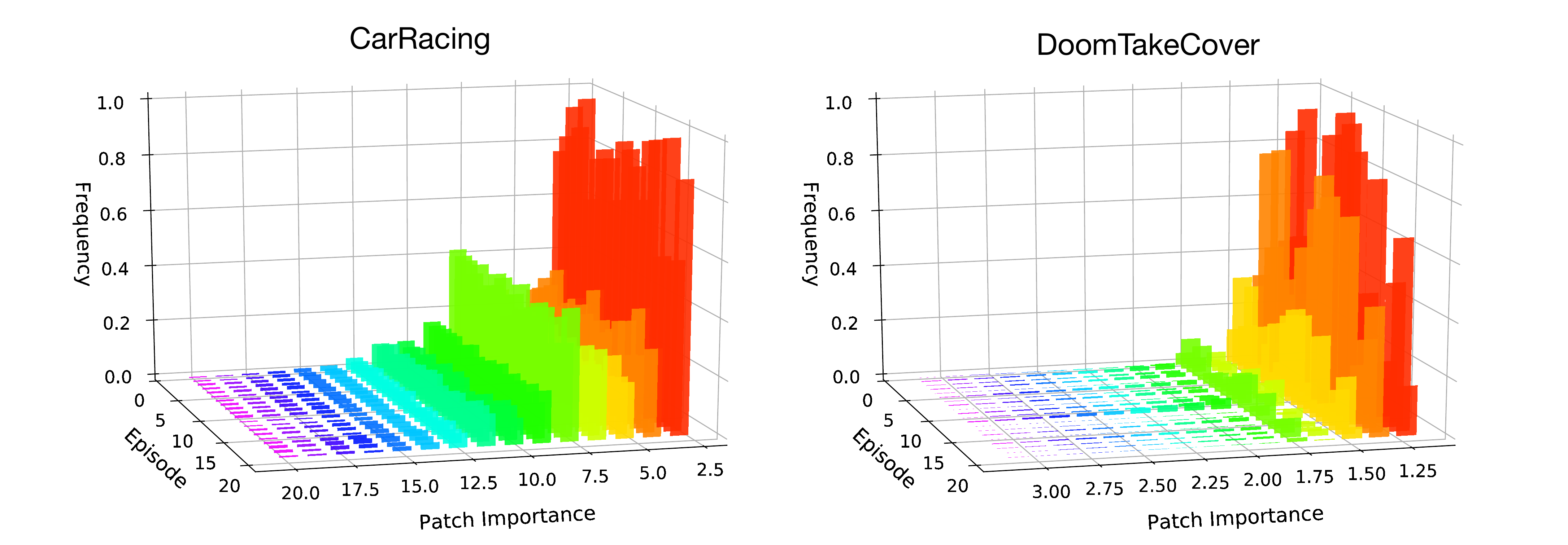}
\vskip -0.05in
\includegraphics[scale=0.22]{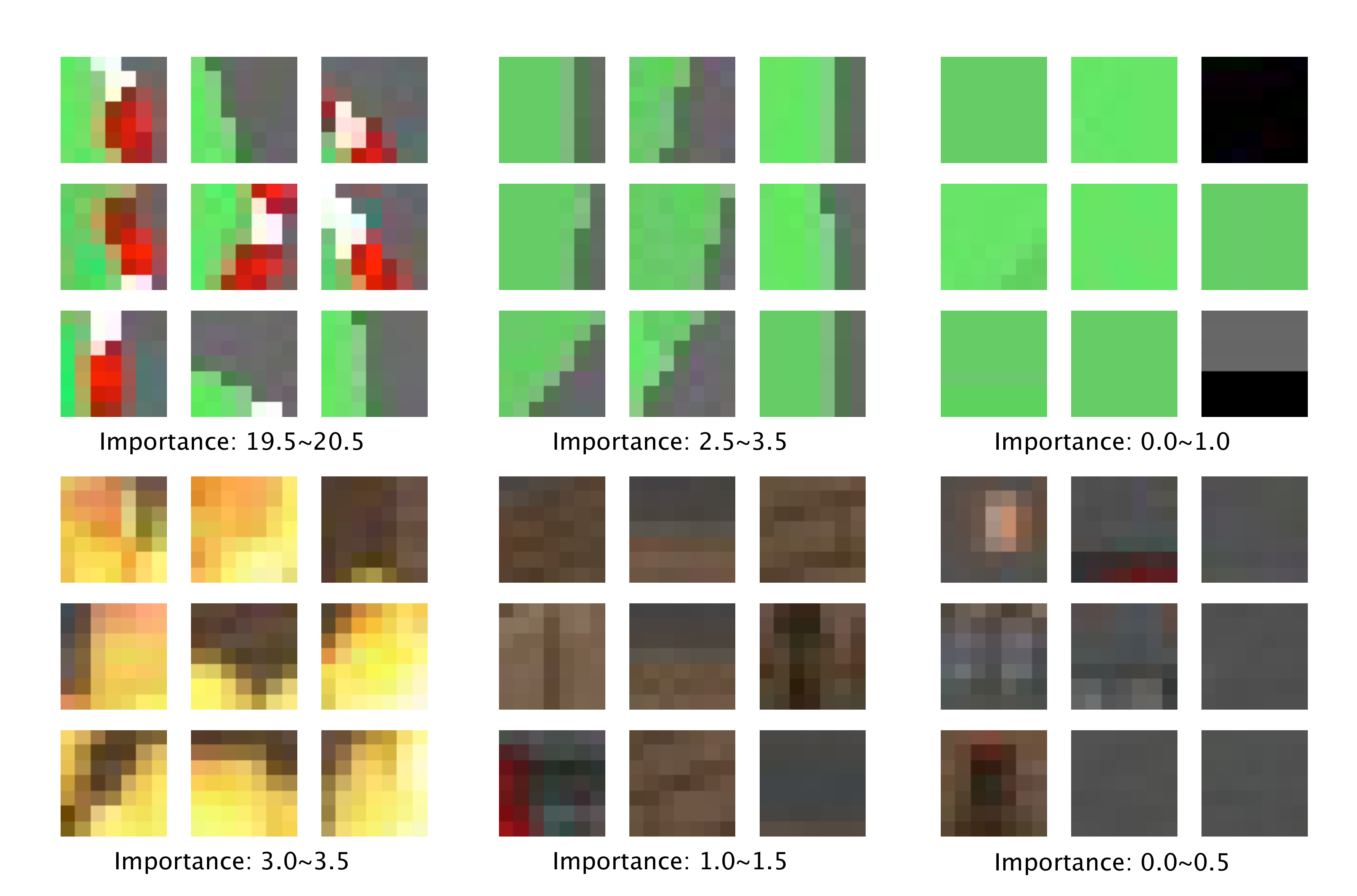}
\vskip -0.15in
\caption{Region of interest to patch importance mapping.
\normalfont Importance voting mechanism via self-attention is able to identify a small minority of patches that are important for the task at hand. The histogram shows the importance distribution of patches from 20 test episodes by patch importance scores (top). Example patches sampled from specified importance ranges (bottom).}
\label{fig:roi_importance_mapping}
\vskip -0.15in
\end{figure}

\subsection{Region of Interest to Importance Mapping}

As our feature retrieval function $f(k)$ is a mapping from patch index to normalized patch center positions, it provides location information, but discards the content in the patches.
On first thought it is surprising to see that the agent is able to solve tasks with the position information alone,
but a closer look at Figure~\ref{fig:roi_importance_mapping} reveals the agent has learned not only \textit{where} but also \textit{what} to attend to.

In Figure~\ref{fig:roi_importance_mapping}, we plot the histogram of patch importance that are in the top $5\%$ quantile from 20 test episodes.
Although each episode presents different environmental randomness controlled by their random seeds at initialization, the distributions of the patch importance are quite consistent, this suggests our agent's behavior is coherent and stable.
When sampling and plotting patches whose importance are in the specified ranges,
we find that the agent is able to map regions of interest (ROI) to higher importance values.
The patches of the highest importance are those critical to the core mission.
These are the patches containing the red and white markers at the turns in CarRacing and the patches having fires in DoomTakeCover (patches on the left).
Shifting to the range that is around the $5\%$ quantile, the patch samples are not as interesting as before but still contains useful information such as the border of the road in CarRacing and the texture of walls in DoomTakeCover.
If we take an extreme and look at the patches with close to zero importance (patches on the right), those patches are mostly featureless and indeed have little information.
By mapping ROIs to importance values, the agent is able to segment and discriminate the input to its controller and learn what the objects are it is attending to.

\subsection{Generalize to Modified Environments}\label{sec:env_modification}

To test our agent's robustness and its ability to generalize to novel states,
we test pre-trained agents in modified CarRacing and DoomTakeCover \textit{without} re-training or fine-tuning.
While there are infinitely many ways to modify an environment,
our modifications respect one important principle: the modifications should not cause changes of the core mission or critical information loss.
With this design principle in mind, we present the following modifications:

\begin{itemize}
    \item \textbf{CarRacing - Color Perturbation} We randomly perturb the background color. At the beginning of each episode, we sample two scalar perturbations uniformly from the interval $[-0.2, 0.2]$ and add respectively to the lane and grass field RGB vectors. Once the perturbation is added, the colors remain constant throughout the episode.
    \item \textbf{CarRacing - Vertical Bars} We add black vertical bars to both sides of the screen. The window size of CarRacing is 800px $\times$ 1000px, and we add two vertical bars of width 75px on the two sides of the window.
    \item \textbf{CarRacing - Background Blob} We add a red blob at a fixed position relative to the car. In CarRacing, as the lane is a closed loop and the car is designed to run in the counter clock-wise direction, the blob is placed to the north east of the car to reduce lane occlusion.
    \item \textbf{DoomTakeCover - Higher Walls} We make the wall higher and keep all other settings  the same.
    \item \textbf{DoomTakeCover - Different Floor Texture} We change the texture of the floor and keep all other settings the same.
    \item \textbf{DoomTakeCover - Hovering Text} We place a blue blob containing text on top part of the screen. The blob is placed to make sure no task critical visual information is occluded.
\end{itemize}

For the purpose of comparison, we used the released code (and pre-trained models, if available) from \cite{ha2018worldmodels,DBLP:conf/gecco/RisiS19} as baselines.
While our reproduced numbers do not exactly match the reported scores, they are within error bounds, and close enough for the purpose of testing for generalization.
For each modification, we test a trained agent for 100 consecutive episodes and report its scores in Table~\ref{table:scores_modified_env}.

\begin{table}[ht]
\vskip -0.05in
\begin{small}
\begin{tabular}{lllll}
\hline
\multicolumn{1}{|l|}{\textbf{CarRacing}} & \multicolumn{1}{l|}{\textbf{Original}} & \multicolumn{1}{l|}{\textbf{Color Perturb}} & \multicolumn{1}{l|}{\textbf{Vertical Bars}} & \multicolumn{1}{l|}{\textbf{Blob}} \\ \hline
\multicolumn{1}{|l|}{WM~\cite{ha2018worldmodels}}     & \multicolumn{1}{l|}{$901 \pm 37$}      & \multicolumn{1}{l|}{$655 \pm 353$}          & \multicolumn{1}{l|}{$\mathbf{166 \pm 137}$} & \multicolumn{1}{l|}{$\mathbf{446 \pm 299}$}   \\ \hline
\multicolumn{1}{|l|}{GA~\cite{DBLP:conf/gecco/RisiS19}}              & \multicolumn{1}{l|}{$859 \pm 79$}      & \multicolumn{1}{l|}{$\mathbf{442 \pm 362}$} & \multicolumn{1}{l|}{$675 \pm 254$} & \multicolumn{1}{l|}{$833 \pm 135$}            \\ \hline
\multicolumn{1}{|l|}{PPO~\cite{xtma_ppo}}              & \multicolumn{1}{l|}{$865 \pm 159$}      & \multicolumn{1}{l|}{$505 \pm 464$} & \multicolumn{1}{l|}{$615 \pm 217$} & \multicolumn{1}{l|}{$855 \pm 172$}            \\ \hline
\multicolumn{1}{|l|}{Ours}            & \multicolumn{1}{l|}{$914 \pm 15$}      & \multicolumn{1}{l|}{$866 \pm 112$}           & \multicolumn{1}{l|}{$900 \pm 35$} & \multicolumn{1}{l|}{$898 \pm 53$}             \\ \hline
                                      &                                        &                                             &                                               \\ \hline
\multicolumn{1}{|l|}{\textbf{TakeCover}} & \multicolumn{1}{l|}{\textbf{Original}} & \multicolumn{1}{l|}{\textbf{Higher Walls}}  & \multicolumn{1}{l|}{\textbf{Floor Texture}} & \multicolumn{1}{l|}{\textbf{Text}}   \\ \hline
\multicolumn{1}{|l|}{WM~\cite{ha2018worldmodels}}     & \multicolumn{1}{l|}{$959 \pm 564$}     & \multicolumn{1}{l|}{$\mathbf{243 \pm 104}$} & \multicolumn{1}{l|}{$\mathbf{218 \pm 69}$} & \multicolumn{1}{l|}{$\mathbf{240 \pm 63}$}    \\ \hline
\multicolumn{1}{|l|}{Ours}            & \multicolumn{1}{l|}{$1125 \pm 589$}    & \multicolumn{1}{l|}{$934 \pm 560$}          & \multicolumn{1}{l|}{$1120 \pm 613$} & \multicolumn{1}{l|}{$1035 \pm 627$}           \\ \hline
\end{tabular}
\caption{Generalization tests.
\normalfont We train agents in the original task and test in the modified environments without re-training. For comparison, we also include the performance in the unmodified tasks. Results with significant performance drop highlighted.}
\label{table:scores_modified_env}
\end{small}
\vskip -0.05in
\end{table}

Our agent generalizes well to all modifications while the baselines fail.
While world model (WM) does not suffer a large performance drop in color perturbations in CarRacing, it is sensitive to all changes.
Specifically, we observe $> 75\%$ score drops in Vertical Frames, Higher Walls, Floor Texture, Hovering Text and a $> 50\%$ score drop in Background Blob from its performances in the unmodified tasks.
Since WM's controller used as input the abstract representations it learned from reconstructing the input images, without much regularization, it is likely that the learned representations will encode visual information that is crucial to image reconstruction but not task critical.
If this visual information to be encoded is modified in the input space, the model produces misleading representations for the controller and we see performance~drop. 

In contrast, GA and PPO performed better at generalization tests. The end-to-end training may have resulted in better task-specific representations learned compared to World model, which uses an unsupervised representation based data collected from random policies. Both GA and PPO can fine-tune their perception layers to assign greater importance to particular regions via weight learning.

Through these tests, we are able to answer the second question in Section~\ref{sec:exp}:
The small change in performance shows that our agent is robust to modifications.
Unlike baseline methods that are subject to visual distractions, our agent focuses only on task critical positions, and simply relies on the coordinates of small patches of its visual input identified via self-attention, and is still able to keep similar performance in the modified tasks~without~any~re-training. By learning to ignore parts of the visual input that it deems irrelevant to the task, it can naturally still perform its task even when irrelevant parts of its environment are modified.

\section{Discussion}

While the method presented is able to cope with various out-of-domain modifications of the environment, there are limitations to this approach, and much more work to be done to further enhance the generalization capabilities of our agent. We highlight some of the limitations of the current approach in this section.

Much of the extra generalization capability is due to attending to the right thing, rather than from logical reasoning. For instance, if we modify the environment by adding a parallel lane next to the true lane, the agent attends to the other lane and drives there instead. Most human drivers do not drive on the opposite lane, unless they travel to another country.

\begin{figure}[htb!]
\vskip -0.05in
\includegraphics[width=0.4775\textwidth]{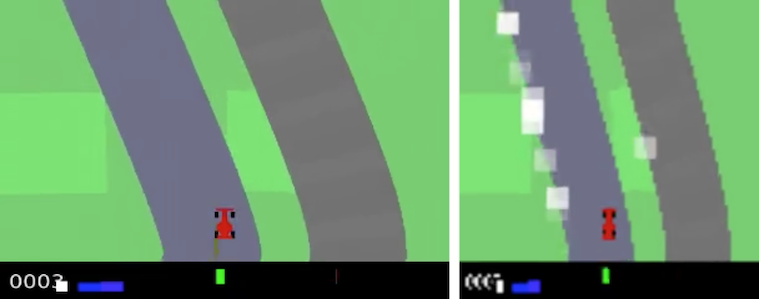}
\vskip -0.10in
\caption{Prefers to drive on the other lane.\;\normalfont When we added an extra fake lane next to the real one, and set the blue channel of the color of the fake lane, the agent attends to the fake lane and proceeds to cross over and drive on it instead of the correct lane.}
\label{fig:carracing_fakelane}
\vskip -0.05in
\end{figure}

We also want to highlight that the visual module does not generalize to cases where dramatic background changes are involved. Inspired by \cite{pineau2018,zhang2018natural}, we modify the background of the car racing environment and replace background with YouTube videos~\cite{catvideo2015}.

\begin{figure}[htb!]
\vskip -0.05in
\includegraphics[width=0.4775\textwidth]{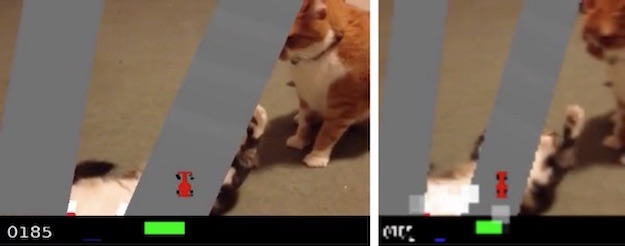}
\vskip -0.05in
\caption{YouTube video background.\;\normalfont The agent stops to look at the cat with the white belly, rather than focus on the road.}
\label{fig:carracing_youtube}
\vskip -0.15in
\end{figure}

\begin{figure}[htb!]
\vskip -0.15in
\includegraphics[width=0.4775\textwidth]{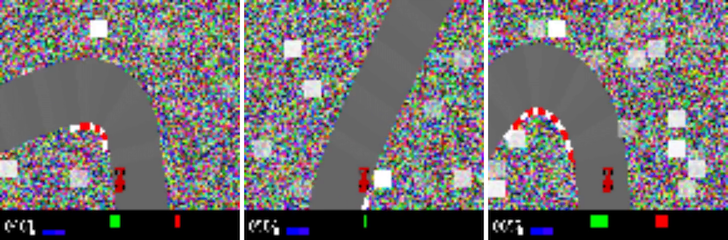}
\vskip -0.05in
\caption{Agent trained from scratch in this noisy environment at various $K$. \normalfont When the background is replaced with random noise, the agent attends only to noise, but never on the road. The policy trained on the normal environment does not transfer to the noisy environment (Score: -58$\pm$12). \textit{Left:} $K=5$ (Score: 452$\pm$210), \textit{Center:} $K=10$ (Score: 577$\pm$243), \textit{Right:} $K=20$ (Score: 660$\pm$259)}
\label{fig:carracing_noise}
\vskip -0.1in
\end{figure}

The agent trained on the original environment with the green grass background fails to generalize when the background is replaced with distracting YouTube videos. When we examine this one step further and replace the background with pure uniform noise, we observe that the agent's attention module breaks down and attends \textit{only} to random noise, rather than to the~road-related~patches. We also experiment with various $K$. When we decrease $K$ from 10 to 5 (or even less), the agent still attends to noisy patches rather than to the road. Not surprisingly, as we increase $K$ to 20 or even 30, the performance of this noise-avoiding policy increases.

These results suggest that while our method is able to generalize to minor modifications, there is much work to be done to approach human-level generalization abilities. The simplistic choice to only use the patch locations (rather than their contents) may be inadequate for more complicated tasks. How we can learn more meaningful features, and perhaps even extract symbolic information from the visual input will be an exciting future direction.

\section{Conclusion}

The paper demonstrated that self-attention is a powerful module for creating RL agents that is capable of solving challenging vision-based tasks.
Our agent achieves competitive results on CarRacing and DoomTakeCover with significantly fewer parameters than conventional methods, and is easily interpretable in pixel space.
Trained with neuroevolution, the agent learned to devote most of its attention to visual hints that are task critical and is therefore able to generalize to environments where task irrelevant elements are modified while conventional methods fail.

Yet, our agent is nowhere close to generalization capabilities of humans. The modifications to the environments in our experiments are catered to attention-based methods. In particular, we have not modified properties of objects of interest, where our method may perform as poorly (or worse) than methods that do not require sparse attention in pixel space. We believe this work complements other approaches (e.g. \cite{kansky2017schema,higgins2017darla,packer2018assessing,zhao2019investigating,agarwal2019learning,hill2019emergent,goyal2019recurrent,Lee2020Network,Song2020Observational,ye2020rotation}) that approach the generalization problem, and future work will continue to develop on these ideas to push the generalization abilities proposed in more general domains~\cite{packer2018assessing,zhang2018natural,cobbe2018quantifying,leibo2018psychlab,juliani2019obstacle,beyret2019animal,cobbe2019leveraging}.

Neuroevolution is a powerful toolbox for training intelligent agents, yet its adoption in RL is limited because its effectiveness when applied to large deep models was not clear until only recently~\cite{such2017deep,DBLP:conf/gecco/RisiS19}.
We find neuroevolution to be ideal for learning agents with self-attention.
It allows us to produce a much smaller model by removing unnecessary complexity needed for gradient-based method.
In addition, it also enables the agent to incorporate modules that include discrete and non-differentiable operations that are helpful for the tasks.
With such small yet capable models, it is exciting to see how neuroevolution trained agents would perform in vision-based tasks that are currently dominated by Deep RL algorithms in the existing literature.

In this work, we also establish the connections between indirect encoding methods and self-attention.
Specifically, we show that self-attention can be viewed as a form of indirect encoding.
Another interesting direction for future works is therefore to explore more specific forms of indirect encoding that, when combined with neuroevolution, can produce RL agents with useful innate behaviors.

\begin{acks}
The authors would like to thank Yingtao Tian, Lana Sinapayen, Shixin Luo, Krzysztof Choromanski, Sherjil Ozair, Ben Poole, Kai Arulkumaran, Eric Jang, Brian Cheung, Kory Mathewson, Ankur Handa, and Jeff Dean for valuable discussions.
\end{acks}

\small

\bibliographystyle{ACM-Reference-Format}
\bibliography{ref} 

\end{document}